\documentclass[runningheads]{llncs}

\usepackage{hyperref}
\usepackage{graphicx}
\usepackage{xcolor}
\usepackage{amsmath,amssymb,amsfonts,mathtools}
\usepackage{algorithmic}
\usepackage{bbm}
\usepackage[linesnumbered,ruled,vlined, french, onelanguage]{algorithm2e}
\SetKwInput{KwInput}{Entrée}     
\SetKwInput{KwOutput}{Sortie}     
\usepackage{multirow}
\usepackage{nccmath}

\title{An Efficient Shapley Value Computation\\
for the Naive Bayes Classifier}
\titlerunning{An Efficient Shapley Value Computation\\
for the Naive Bayes Classifier}

\author{Vincent Lemaire \and
Fabrice Cl\'erot \and
Marc Boull\'e}

\authorrunning{V. Lemaire et al.}

\institute{Orange Innovation, Lannion, France}

\begin{document}

\maketitle

\begin{abstract}
Variable selection or importance measurement of input variables to a machine learning model has become the focus of much research. It is no longer enough to have a good model, one also must explain its decisions. This is why there are so many intelligibility algorithms available today. Among them, Shapley value estimation algorithms
are intelligibility methods based on cooperative game theory.
In the case of the naive Bayes classifier, and to our knowledge, there is no ``analytical" formulation of Shapley values. This article proposes an exact analytic expression of Shapley values in the special case of the naive Bayes Classifier. We analytically compare this Shapley proposal, to another frequently used indicator, the Weight of Evidence (WoE) and provide an empirical comparison of our proposal with (i) the WoE and (ii) KernelShap results on real world datasets, discussing similar and dissimilar results. The results show that our Shapley proposal for the naive Bayes classifier provides informative results with low algorithmic complexity so that it can be used on very large datasets with extremely low computation time. 

\end{abstract}

%-------------------------------------------------------
\section{Introduction}
%-------------------------------------------------------

There are many intelligibility algorithms based on the computation of variable's contribution to classifier results, often empirical and sometimes without theoretical justifications. This is one of the main reasons why the Python SHAP library was created in 2017 by Scott Lundberg following his publication \cite{Lundberg2017}, to provide algorithms for estimating Shapley values, an intelligibility method based on cooperative game theory. Since its inception, this library has enjoyed increasing success, including better theoretical justifications and qualitative visualizations. It provides local explanation like other methods such as LIME \cite{lime}.

In the case of the naive Bayes classifier, we show in this paper that Shapley values can be computed accurately and efficiently. The key contributions are:

\begin{itemize}
    \item an analytical formula for the Shapley values in the case of the naive Bayes classifier,
    \item an efficient algorithm for calculating these values, with algorithmic complexity linear with respect to the number of variables.
\end{itemize}

The remainder of this paper is organized into three contributions : (i) in the next section \ref{shapley} we give our proposal for local Shapley values in the case of the naive Bayes (NB) classifier, with further discussion in the section \ref{discussion}; (ii) the following section \ref{comparewoe} compares, in an analytic analysis, our Shapley proposal to another frequently used indicator in the case of the NB classifier: the Weight of Evidence (WoE); (iii) we then provide, in section \ref{expe} an empirical comparison of the results our Shapley formulation to the results of (i) the WoE and (ii) KernelShap on real world datasets and discuss similar similar and dissimilar results. The last section concludes the paper.

%-------------------------------------------------------
\section{Shapley for naive Bayes Classifier}
\label{shapley}

To our knowledge, there is no ``analytical" formula of Shapley values for the naive Bayes classifier in the literature\footnote{See the introduction of the Section 4 for a very brief literature overview}. This first section is therefore devoted to a proposal for calculating these these values,  exploiting the conditional variable independence assumption that characterizes this classifier .

\subsection{Reminders on the naive Bayes classifier}
\label{snb} 

The naive Bayes classifier (NB) is a widely used tool in supervised classification problems. It has the advantage of being efficient for many real data sets \cite{Hand2001}. However, the naive assumption of conditional independence of the variables  can, in some cases, degrade the classifier's performance. This is why variable selection methods have been developed \cite{Langley1994}. They mainly consist of variable addition and deletion heuristics to select the best subset of variables maximizing a classifier performance criterion, using a wrapper-type approach \cite{Guyon2003}. It has been shown in \cite{BoulleJMLR07} that averaging a large number of selective naive Bayes classifiers\footnote{In this case, it is an assembly of models providing better results than a single classifier}, performed with different subsets of variables, amounts to considering only one model with a weighting on the variables. Bayes' formula under the assumption of independence of the input variables conditionally to the class variable becomes:

\begin{equation}
\medmath{
P(Y_{k}|X)=\frac{P(Y_{k})\prod_{i}P(X_{i}|Y_{k})^{w_i}}{\sum_{j=1}^{K}(P(Y_{j})\prod_{i}P(X_{i}|Y_{j})^{w_i})}}
\label{snb2}
\end{equation}

where $w_i \in [0, 1]$ is the weight of variable $i$.
The predicted class is the one that maximizes the conditional probability $P(Y_{k}|X)$. The probabilities $P(X_i|Y_i)$ can be estimated by interval using discretization for numerical variables. Gaussian naive Bayes could be also considered. For categorical variables, this estimation can be done directly if the variable takes few different modalities, or after grouping (of values) in the opposite case. 

Note 1: in accordance with the naive Bayes model definition, our Shapley value proposal assumes that the variables of the model are independent conditionally to the class. In practice, we expect a variable selection method to result in a classifier relying on variables which are uncorrelated or only weakly correlated conditionally to the class. A posthoc analysis of our results shows that this is indeed the case in the experiments of this article with the parsimonious classifier used (see Section \ref{classifier}).

Note 2: Even if in equation \ref{snb2} the NB have transparent weights for all feature variables it is interesting to explain NB models in order to have local interpretations.

\subsection{Definition and notations}

The following notations are used:
\begin{itemize}
\item[*] the classifier uses $d$ variables: $[d] = \{1, 2, ..., d\}$
\item[*] for a subset $u$ of $[d]$, we note $|u|$ the cardinality of $u$
\item[*] for two disjoint sets $u$ and $r$ of $[d]$, let $u+r$ be $ u \cup r $
\item[*] for a subset $u$ of $[d]$, we denote by {\small $-u = [d]\backslash u$}, the complement of $u$ in $d$
%item ... ?
\end{itemize}

We define a ``value function'' $v(.)$ indicating for each subset $u$ of variables the maximum ''contribution'' they can obtain together, i.e. $v(u)$, to the output of the classifier. The maximum value (or total gain) of the value function is reached when all the variables are taken into account, $v([d])$. The Shapley value for variable $j$ is denoted $\phi_j$.
Shapley's theorem \cite{shapley_shubik_1954} tells us that there is a unique distribution of Shapley values satisfying the following four properties:
\begin{itemize}
\item Efficiency: $v([d])=\sum_j \phi_j$; i.e. the total gain is distributed over all the variables
\item Symmetry:  if $\forall u \subset -\{i,j\}$, $v(u+j)=v(u+i)$, then $\phi_j=\phi_i$; i.e. if the variables $i$ and $j$ bring the same gain to any subset of variables, then they have the same Shapley value
\item Null player: if $\forall u \subset -\{i\}$, $v(u+i)=v(u)$, then $\phi_i=0$; i.e.  if the variable $i$ contributes nothing to any subset of variables, then its Shapley value is zero

\item Additivity: if the $d$ variables are used for two independent classification problems $A$ and $B$ associated with $v_A, v_B$, then the Shapley values for the set of two problems are the sum of the Shapley values for each problem
\end{itemize}

\subsection{Shapley Values for the naive Bayes Classifier}

{\bf 2.3.1 `Value Function': } In the case of the NB we propose to take as `Value Function' (case of a two-class classification problem) the log ratio (LR) of probabilities:

\begin{eqnarray}
\medmath{LR} & = & \medmath{log \left( \frac{P(Y_{1}|X)}{P(Y_{0}|X)} \right)} \nonumber \\
 & = & \medmath{log\left( \frac{P(Y_{1})\prod_{i=1}^{d}P(X_{i}|Y_{1})^{w_i}}{\sum_{j=1}^{K}(P(Y_{j})\prod_{i=1}^{d}P(X_{i}|Y_{j})^{w_i})}  \frac{\sum_{j=1}^{K}(P(Y_{j})\prod_{i=1}^{d}P(X_{i}|Y_{j})^{w_i})} {P(Y_{0})\prod_{i=1}^{d}P(X_{i}|Y_{1})^{w_i}} \right)} \nonumber \\
& = & \medmath{log\left( \frac{P(Y_{1})\prod_{i=1}^{d}P(X_{i}|Y_{1})^{w_i}}{P(Y_{0})\prod_{i=1}^{d}P(X_{i}|Y_{1})^{w_i}} \right)} \nonumber \\
& = & \medmath{log\left( \frac{P(Y_{1})}{P(Y_{0})}\right) + \sum_{i=1}^{d}
{ w_i log \left( \frac{P(X_{i}|Y_{1})}{P(X_{i}|Y_{0})}  \right)    }}
\label{LR5}
\end{eqnarray}

The choice of the logarithm of the odd ratio as the "value function" is motivated by two reasons (i) the logarithm of the odd ratio is in bijection with the score produced by the classifier according to a monotonic transformation (ii) the logarithm of the odd ratio has a linear form that allows the derivation of an analytical formula. This value function differs from the usual value function, $f(X)=P(Y|X)$, as mentioned and analyzed later in this document when comparing with KernelShap (see section \ref{expeshap}).

We stress here that the derivation above is only valid in the case of independent variables conditionally to the class variable, which is the standard assumption for the naive Bayes classifier.

For a subset, $u$, of the variables \footnote{on the covariates in $u$, we average over the conditional distribution of $X_{-u}$} given $X_u = x_u$:
\begin{equation}
v(u) = \mathbb{E}_{X_{-u} | X_u = x_u} \left[ LR(X_u = x_u^*, X_{-u}) \right]
\end{equation}
which we write in a "simplified" way afterwards
\begin{equation}
v(u)=\mathbb{E} \left[ (LR(X)|X_u=x_u^*) \right]
\label{simple}
\end{equation}
This is a proxy of the target information provided by $u$ at the point $X = x^*$.
Thus, for a point (an example) of interest $x^*$ we have:
\begin{itemize}
\item $v([d])=LR(X=x^*)$,  everything is conditional on $x^*$ so we have the log odd ratio for $X=x^*$
\item $v(\emptyset)=\mathbb{E}_X \left[ LR(X) \right] =\mathbb{E}_X \left[ log (\frac{P(Y_1|X)}{P(Y_0|X)} ) \right]$,  nothing is conditioned so we have the expectation of the log odd ratio
\end{itemize}

{\bf 2.3.2 Shapley Values: } By definition of the Shapley values \cite{shapley_shubik_1954}, we have for a variable $m$:
\begin{equation}
\phi_m=\frac{1}{d} \sum_{u \in -m} \frac{v(u+m)-v(u)}{ \binom{d-1}{|u|}}
\label{sumshap}
\end{equation}
To obtain $\phi_{m}$, we therefore need to calculate, for a subset of variables in which the variable $m$ does not appear, the difference in gain $v(u+m)-v(u)$. This makes it possible to compare the gain obtained by the subset of variables with and without the $m$ variable, in order to measure its impact when it "collaborates" with the others.

We therefore need to calculate $v(u+m)-v(u)$ in the case of the naive Bayes classifier.
If this difference is positive, it means that the variable contributes positively. Conversely, if the difference is negative, the variable is penalizing the gain.
Finally, if the difference is zero, it indicates that the variable makes no contribution. Following the example of \cite{Lundberg2017} and Corollary1 with a linear model whose covariates are the log odd ratio as a `value function', one can decompose the subsets of variables into 3 groups $ \{u\}, \{m\}, -\{u+m\}$.

{\bf Calculation} of $v(u)$ : On $\{u\}$,  we condition on $X_u = x_u$  while on $\{m\}$, $\{u+m\}$, we do an averaging. 
By consequent:
\begin{eqnarray} 
v(u)  & = &  \medmath{ \mathbb{E}\left[ LR(X) | X_u = x_u^*) \right] } \\ \nonumber
        & = & \medmath{ log(P(Y_1)/P(Y_0)) } \\ \nonumber
        & + & \medmath{ \sum_{k \in u} w_k log\left( \frac{P( X_k={x_k}^* | Y_1)}{P( X_k={x_k}^* | Y_0)} \right) } \\ \nonumber
        & + & \medmath{ w_m \sum_{X_m} \left[  P(X_m=x_m) log\left( \frac{P( X_m=x_m | Y_1)}{P( X_m=x_m | Y_0)} \right) \right] }\\ \nonumber
        & + & \medmath{ \sum_{k \in -\{u+m\}} w_k \sum_{X_k} \left[ P(X_k=x_k) log\left( \frac{P( X_k={x_k} | Y_1)}{P( X_k={x_k} | Y_0)} ) \right)\right] }\\
\end{eqnarray}

{\bf Calculation of $v(u+m)$} : The only difference is that we also condition on $X_m$ 
\begin{eqnarray} 
v(u+m)  & = &  \medmath{ \mathbb{E}\left[ LR(X) | X_{u+m} = x_{u+m}^*) \right] }\\ \nonumber
        & = & \medmath{ log(P(Y_1)/P(Y_0)) } \\ \nonumber
        & + & \medmath{ \sum_{k \in u} w_k log\left( \frac{P( X_k={x_k}^* | Y_1)}{P( X_k={x_k}^* | Y_0)} \right) } \\ \nonumber
        & + & \medmath{ w_m  \left[ log\left( \frac{P( X_m=x_m^* | Y_1)}{P( X_m=x_m^* | Y_0)} \right) \right] } \\ \nonumber
        & + & \medmath{ \sum_{k \in -\{u+m\}} w_k \sum_{X_k} \left[ P(X_k=x_k) log\left( \frac{P( X_k={x_k}^* | Y_1)}{P( X_k={x_k}^* | Y_0)} ) \right)\right] }\\
\end{eqnarray}

The difference $v(u+m) - v(u)$ is independent on $u$ and therefore the combinatorial sum averaging over all $u \in -m$ in equation \ref{sumshap} simply vanishes and finally $\phi_m=v(u+m) - v(u)$
\begin{eqnarray}
&=& \medmath{w_m \left(  log\left( \frac{P( X_m=x_m^* | Y_1)}{P( X_m=x_m^* | Y_0)} \right)  -   \sum_{X_m} \left[   P(X_m=x_m) log\left(  \frac{P( X_m=x_m | Y_1)}{P( X_m=x_m | Y_0)}  \right)  \right]  \right)}\nonumber \\
&=& \medmath{w_m \left(  log\left( \frac{P( X_m=x_m^* | Y_1)}{P( X_m=x_m^* | Y_0)} \right)  -   \mathbb{E} \left( log\left(  \frac{P( X_m=x_m | Y_1)}{P( X_m=x_m | Y_0)}  \right)  \right)  \right)}
 \label{eqxx}
\end{eqnarray}

Equation \ref{eqxx} provides the exact analytical expression of the Shapley value for our choice of the log odd ratio as the value function of the weighted naive Bayes.\\

\section{Interpretation and Discussion}
\label{discussion} 

We give here some interpretation details and discussion about the Shapley formulation which are interesting arguments for its use.

 $\bullet$ The equation \ref{eqxx} is the difference between the information content of $X_m$ conditionally on $X_m = x_m^*$ and the expectation of this information. In other words, it is the information contribution of the variable $X_m$ for the value $X_m = x_m^*$ of the considered instance, contrasted by the average contribution on the entire database.

 $\bullet$ The Equation \ref{eqxx} can be rewritten (we just omit the product by $w_m$) in the form: 
\begin{eqnarray}
&   & \medmath{- \left[ log \left( \frac{1}{P(X_m=x_m^{*}|Y_1)} \right) - \sum_{X_m} \left( P(X_m=x_m) log \left(  \frac{1}{P( X_m=x_m | Y_1)}  \right) \right) \right]} \nonumber  \\ 
& &  \medmath{+ \left[ log \left( \frac{1}{P(X_m=x_m^*|Y_0)} \right) - \sum_{X_m} \left(   P(X_m=x_m) log\left(  \frac{1}{P( X_m=x_m | Y_0)}  \right)  \right)\right]}  \nonumber \\
\label{eqyy}
\end{eqnarray}

The terms in brackets […] in equation \ref{eqyy} are the difference between the information content related to the conditioning $X_m = x_m^*$ and the entropy of the variable $X_m$ for each class ($Y_0$ and $Y_1$). This term measures how much conditioning on $X_m = x_m^*$ brings information about the target classes.\\

$\bullet$ For a given variable, the expectation of our Shapley proposal is equal to zero, due to the conditional independence of the variables. The consequence is that high
Shapley values in some parts of the data space must be exactly
compensated by low values in other parts of the data space.

$\bullet$ For a given example if we return to our choice of value function (equation~\ref{LR5}) and using the sum of equation~\ref{eqxx} over the $d$ variables
we have:

\begin{eqnarray}
LR & = & \medmath{log\left( \frac{P(Y_{1})}{P(Y_{0})}\right) + \sum_{m=1}^{d}
{ w_m log \left( \frac{P(X_{m}|Y_{1})}{P(X_{m}|Y_{0})}  \right)    }} \nonumber \\
&=& \medmath{log\left( \frac{P(Y_{1})}{P(Y_{0})}\right) +  \sum_{m=1}^{d} \phi_m + \sum_{m=1}^{d} \mathbb{E} \left( log\left(  \frac{P( X_m=x_m | Y_1)}{P( X_m=x_m | Y_0)}  \right)  \right) } \nonumber \\
&=&   \medmath{\sum_{m=1}^{d} \phi_m + \mbox{cste} \label{eq15}} 
\end{eqnarray}

We obtain a result consistent  with the notion of a value function for the Shapley's formulation. Our value function consists of a constant plus the individual contribution of the $d$ variables. The constant is the log ratio of class prior plus the sum of the average contribution of all variables.

$\bullet$ If we inverse the role of $Y_0$ and $Y_1$ in equation \ref{eqxx}, we observe that the Shapley value is symmetric; i.e the positive contribution of the variable for $Y_0$ is negative for $Y_1$ (with the same absolute value).

$\bullet$  When the numerical (resp. categorical) variables have been previously discretized into intervals (resp. groups of values), the complexity of the equation \ref{eqxx} is linear in the number of discretized parts. For an input vector made up of $d$ variables, this complexity is $O(\sum_{i=1}^d P_i)$ where $P_i$ is the number of discretized parts of variable $i$. 

$\bullet$ In term of explainability, if the discretization method used for numerical attributes (resp. grouping method for categorical attributes) provides a reasonable number of intervals (resp. groups of values), then the number of potential ``behaviors" of the individuals in the classification problem is small and therefore easy to understand.

 $\bullet$ Extension to multiclass: We simply define the Shapley Value of an input variable as the sum of the absolute $C$ Shapley values when choosing in equation~\ref{eqxx} one of the $C$ class of the problem as the ``positive class'' ($Y_1$)  and all the others $C-1$ class as the ``negative class''  ($Y_0$).
For example in a 3 class problems where the class are `red', 'green', and 'yellow':
\begin{eqnarray}
\phi_m & =  \medmath{  | \phi_{m}(Y_1=\{red\},Y_0=\{green,yellow\}) | }\nonumber \\ 
& \medmath{ +  | \phi_{m}(Y_1=\{green\},Y_0=\{red,yellow\}) | } \nonumber \\ 
& \medmath{ +  | \phi_{m}(Y_1=\{yellow\},Y_0=\{green,red\}) | } \nonumber 
\end{eqnarray}
In this way, we can find out which feature has the greatest impact on all classes. Note that there are other ways of measuring the impact of features in multi-classification problems (see, for example, the discussion in \cite{discussion} on using the SHAP package for multi-classification problems).

\begin{figure}[t]
    \centering
    \includegraphics[width=0.7\linewidth]{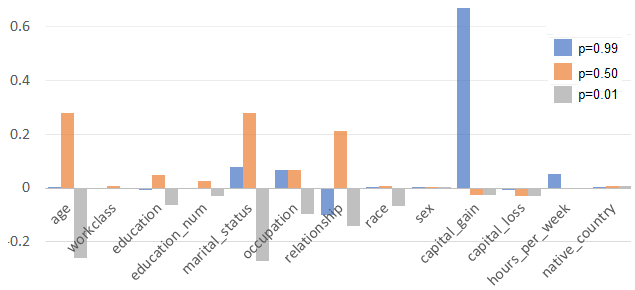}
    \caption{Normalized Shapley values - Illustrative example ($\frac{\phi_m}{\sum_{i=1}^{d}\phi_i}$)}
    \label{fig:illustrative}
\end{figure}

 $\bullet$ To conclude this discussion and prior to the experiments presented in Section~\ref{expe}, we give here an illustrative example on the Adult dataset (the experimental conditions are the same as those presented in Section~\ref{expe}). Figure~\ref{fig:illustrative} shows the Shapley values obtained for 3 examples which are respectively predicted as belonging to the class `more' with probabilities 0.99, 0.50 and 0.01. On this well-known data set, we find the usual results on the role of input variables for examples with high to low probabilities when considering the class `more'.

\section{Analytic comparison with the Weight of Evidence}
\label{comparewoe}

In the case of the naive Bayes classifier,  there are a number of ``usual'' methods for calculating the importance of input variables. We do not go into detail on all of them, but the reader can find a wide range of these indicators in \cite{Robnik2008,LemaireIJCNN2010amethod} for a brief literature overview but nonetheless quite exhaustive. This section focuses on presenting  the ``Weight of evidence" (WoE) \cite{good1950} and its comparison with the Shapley values proposed in the previous section, since this indicator is (i) close to the equation presented above (equation \ref{eqxx}) and (ii) among the most widely used indicators for the naive Bayes classifier.

We give below the definition of the WoE (in the case with two classes) which is a log odds ratio calculated between the probability of the output of the model and the latter deprived of the variable $X_m$:
\begin{equation}
\medmath{(WoE)_m}=\medmath{log \left( \frac{\frac{p}{1-p}}{\frac{q}{1-q}} \right)} = \medmath{w_m  \left( log \left( \frac{\frac{P(Y_1|X)}{P(Y_0|X)}}{\frac{P(Y_1|X \backslash X_m)}{P(Y_0|X \backslash X_m)}} \right)  \right)}=  \medmath{w_m  \left( log \left( \frac{P(Y_1|X) P(Y_0|X \backslash X_m)}{P(Y_0|X) (Y_1|X \backslash X_m)} \right)  \right) } \label{tutotu}
\end{equation}
\begin{eqnarray}
\medmath{(WoE)_m}& =&  \medmath{w_m  \left( log \left( \frac{P(Y_1) \left[ \prod_{i=1}^{d} P(X_i|Y_1) \right] P(Y_0) \left[ \prod_{i=1, i\ne m}^{d} P(X_i|Y_0) \right] }{P(Y_0) \left[ \prod_{i=1}^{d} P(X_i|Y_0) \right] P(Y_1)  \left[ \prod_{i=1, i\ne m}^{d} P(X_i|Y_1) \right]} \right) \right)}
\label{tututu}
\end{eqnarray}
by simplifying the numerator and denominator:
\begin{equation}
\medmath{(WoE)_m = w_m \left( 
 log\left( \frac{P( X_m=x_m^* | Y_1)}{P( X_m=x_m^* | Y_0)} \right)  \right)}
 \label{eq18}
\end{equation}

{\bf Link between $(WoE)_m$ and $\phi_m$}: 
If we compare the equations \ref{eq18} and \ref{eqxx}, we can see that it is the reference that changes. For the Shapley value ( equation \ref{eqxx}), the second term takes the whole population as a reference whereas  for the WoE (equation \ref{eq18}) the reference is zero. The averaging is not at the same place between the two indicators, as we will demonstrate just below. We can also observe that the expectation of our Shapley proposal is equal to zero, whereas the expectation of WoE is the second term of our Shapley proposal (second part, the expectation term, of equation \ref{eqxx}).

In case of the naive Bayes classifier, ``depriving" the classifier of a variable is equivalent to performing a ``saliency" calculation (as proposed in \cite{LemaireBook2006input})  which takes into account the probability distribution of the variable $X_m$. Indeed, to deprive the classifier of the variable $X_m$, it is sufficient to recalculate the average of the classifier's predictions for all the possible values of the variable $X_m$ as demonstrated in \cite{Robnik2008}.
Indeed, if we assume that the variable $X_m$ has $k$ distinct values, Robnik et al. \cite{Robnik2008} have shown that the saliency calculation of \cite{LemaireBook2006input} is exact in the naive Bayes case and amounts to ``erasing" the variable $X_m$. Denoting either $Y = Y_0$ or $Y = Y_1$ by $Y_.$, we have

\begin{eqnarray}
\medmath{ P(Y_.|X \backslash X_m) } & =& \sum_{q=1}^{k} P(X_m=X_q) \frac{P(Y_.|X , X_m=X_q)}{P(X , X_m=X_q)}\\
\medmath{ P(Y_.|X \backslash X_m) } & =&  \sum_{q=1}^{k} P(X_m=X_q) \medmath{ \left( P(Y_.) \left( \prod_{i=1, i\ne m}^{d} \frac{P(X_i|Y_.)}{P(X_i)} \right) \frac{P(X_m=X_q|Y_.)}
{P(X_m=X_q)} \right) } \nonumber 
\end{eqnarray}
\begin{eqnarray}
\medmath{ P(Y_.|X \backslash X_m) } & =& P(Y_.) \prod_{i=1, i\ne m}^{d} P(X_i|Y_.) \medmath{\left( \sum_{q=1}^{k} \frac{P(X_m=X_q) P(X_m=X_q|Y_.)}{P(X_m=X_q)} \right)} \label{tatata} \\
\medmath{ P(Y_.|X \backslash X_m) } & =& \medmath{ P(Y_.) \prod_{i=1, i\ne m}^{d} P(X_i|Y_.) }
\end{eqnarray}

with $P(Y_.|X , X_m=X_q)$ being $P(Y_.|X)$ but where the value of the variable $X_m$ has been replaced by another value of its distribution $X_q$.
This last result is interesting because with the help of the equation \ref{tatata} we can rewrite the equation \ref{tutotu} in :
\begin{eqnarray}
%\scriptsize
\medmath{(WoE)_m} & =&  w_m  \left( \medmath{ log \left( \frac{P(Y_1|X) P(Y_0|X \backslash X_m)}{P(Y_0|X) (Y_1|X \backslash X_m)} \right) } \right)  \nonumber \\
\medmath{(WoE)_m} & =&  w_m   log \frac{\left( \medmath{P(Y_1) \prod_{i=1}^{d} P(X_i|Y_1)}\right) \medmath{\left( P(Y_0) \prod_{i=1, i\ne m}^{d} P(X_i|Y_0) \sum_{q=1}^{k} P(X_m=X_q|Y_0) \right)} }{ \left( \medmath{P(Y_0) \prod_{i=1}^{d} P(X_i|Y_0)} \right) \medmath{\left( P(Y_1) \prod_{i=1, i\ne m}^{d} P(X_i|Y_1) \sum_{q=1}^{k} P(X_m=X_q|Y_1) \right)} }    \nonumber \\
\medmath{(WoE)_m} &=& w_m \left( \medmath{ log\left( \frac{P( X_m=x_m^* | Y_1)}{P( X_m=x_m^* | Y_0)} \frac{\sum_{q=1}^{k} P(X_m=X_q|Y_0)}{\sum_{q=1}^{k} P(X_m=X_q|Y_1)} \right)}  \right) \nonumber \\
\medmath{(WoE)_m} &=& w_m \left( \medmath{ log\left( \frac{P( X_m=x_m^* | Y_1)}{P( X_m=x_m^* | Y_0)} \right)}   +  \medmath{ log\left( \frac{\sum_{q=1}^{k} P(X_m=X_q|Y_0)}{\sum_{q=1}^{k} P(X_m=X_q|Y_1)} \right)}  \right) \label{eq22} \\
\medmath{(WoE)_m} &=& w_m \left( \medmath{ log\left( \frac{P( X_m=x_m^* | Y_1)}{P( X_m=x_m^* | Y_0)} \right) }  +  \medmath{log\left( \frac{1}{1} \right) } \right) \label{tititi}
\end{eqnarray}

This result allows to better understand why the WoE is referenced in zero. The comparison of the equation \ref{eqxx} and the equation \ref{eq22} exhibits the difference in the localization of the averaging resulting in a reference in zero for the WoE. In the first case an expectation is computed on the variation of the log ratio $\medmath{log(P(Y_1|X)/P(Y_0|X))}$ while in the second case this expectation is computed only on the variations of $f(X)=P(Y_1|X)$ (or reciprocally $P(Y_0|X)$). 

This comparison shows the effect of choosing either the odds (our Shapley proposal) or the output of the classifier (WoE) as the `value function'. Since both results are very consistent, and WoE does not suffer from calculation exhaustion, the two methods are very close.

%%%%%%%%%%
\section{Experiments}
\label{expe}

The experiments carried out in this section allow us to  compare our Shapley proposal with the Weight of Evidence and KernelShap to highlight similar or dissimilar behaviors. We focus below on two classes problems. 

The code and data used in this section are available in the  GitHub repository at \url{https://tinyurl.com/ycxzkffk}.

\subsection{Datasets and Classifier}
\label{classifier}
{\bf Classifier} : The naive Bayes classifier used in the experiments exploits two main steps. A first step in which  (i) the numerical variables are discretized, 
using the method described in \cite{BoulleML06},
(ii) the modalities of the categorical variables are grouped
using the method described in \cite{BoulleJMLR05}. 
Then, variable weights are calculated using the method described in \cite{BoulleJMLR07}. In the first and second steps, uninformative variables are eliminated from the learning process. In this paper, we have used the free Khiops software  \cite{khiops} in which  the whole process is implemented. This software produces a preparation report containing a table of the values of $P(X_m=x_m|Y_.)$ for all classes and all variables, enabling us to easily implement the two methods described earlier in the article. 

Note: below, the same classifier and preprocessing are used for comparing the different methods used to calculate the variable importance, so that the differences in the results will be only due to those different methods.

{\bf Dataset} : Ten datasets have been selected in this paper and  are described in the Table~\ref{table:dataset}. They are all available on the UCI website  \cite{Lichman2013} or on the Kaggle website\cite{kaggle}. They were chosen to be representative datasets in terms of variety of number of numerical attributes (\#Cont), number of categorical attributes (\#Cat), number of instances (\#Inst) and imbalance between classes\footnote{Here we give the percentage of the majority class.} (Maj. class.). They are widely used in the ``machine learning" community as well as  in the analysis of recently published Shapley value results. 
In this table, we give in the last columns the performances, for information purposes, obtained by the  naive Bayes used (an averaged naive Bayes, see Section \ref{snb}); i.e the accuracy and the Area Under the ROC curve (AUC), as well as the number of variables retained by this classifier (\#Var) since uninformative variables are eliminated from the learning process. As the aim of this article is not to compare classification results, we decide simply to use 100 \% of the examples to train the model\footnote{To facilitate reproducibility. Nevertheless, the test performances of the models (Table \ref{table:dataset})  are very close with a 10-fold cross-validation process.}  and to compute later the  importance indicators (WoE and Shapley) .

\begin{table}[htbp]
\centering
\fontsize{9}{8}\selectfont
\begin{tabular}{|l|c|c|c|c|c|c|c|c|}
 \hline
Name & \#Cont & \#Cat & \#Inst ($N$)  & Maj. class. & Accuracy & AUC & \#Var\\ \hline 
Twonorm   & 20 & 0 & 7400  & 0.5004 & 0.9766 & 0.9969 & 20\\
Crx   & 6 & 9 & 690 & 0.5550 & 0.8112 & 0.9149 & 7\\
Ionosphere   & 34 & 0 & 351 & 0.6410 & 0.9619 & 0.9621 & 9 \\
Spam   & 57 & 0 & 4307 & 0.6473 & 0.9328 & 0.9791 & 29 \\
Tictactoe   & 0 & 9 & 958 & 0.6534 & 0.6713 & 0.7383 & 5\\
German   & 24 & 0 & 1000 & 0.7 & 0.7090 & 0.7112 & 9 \\
Telco   & 3 & 18 & 7043 & 0.7346 & 0.8047 & 0.8476 & 10 \\
Adult   & 7 & 8 & 48842 & 0.7607 & 0.8657 & 0.9216 & 13 \\
KRFCC   & 28 & 7 & 858 & 0.9358 & 0.9471 & 0.8702 & 3\\
Breast   & 10 & 0 & 699 & 0.9421 & 0.975 & 0.9915 & 8\\
\hline
\end{tabular}
\caption{Description of the datasets used in the experiments (KRFCC = KagRiskFactorsCervicalCancer dataset)}
\label{table:dataset}
\end{table}

\subsection{Comparison with the WoE}
\label{comparisonwithwoe}

In this first part of the experiments, the comparison is made with the Weight of Evidence. and we present  the  observed correlation between the Shapley values (Eq. \ref{eqxx}) and the WoE values (Eq. \ref{eq18}). 

We compute the Shapley and WoE values per class ($C$), per variable ($J$) and  per instance ($N$) then, we compute the Kendall correlation\footnote{We used the scipy.stats.kendalltau with the default parameter, i.e $\tau$-b.} line per line;  that is, for each example, we compute the $d$ values of WoE or of our Shapley values and then the Kendall coefficient for that example. Finally we compute the average and the standard deviation of these $N$ values  which are reported in the Table \ref{table:corr-woe-2C}. 

The Kendall correlation is a measure of rank correlation, therefore, it measures whether the two indicators, WoE and our Shapley values, give the same ordering in the importance of the variables.

\begin{table}[!ht]
\centering
%\fontsize{9}{8}\selectfont
\begin{tabular}{|l|c|c|}
 \hline
Name & Kendall \\\hline 
Twonorm	    &  0.9919 $\pm$8.71e-05\\
Crx	        &  0.9919 $\pm$4.28e-04\\
Ionosphere	&  0.8213 $\pm$1.76e-02\\
Spam	    &  0.9011 $\pm$2.66e-04\\
Tictactoe	&  1.0000 $\pm$2.60e-04\\
German	    &  0.9515 $\pm$1.01e-03\\
Telco	    &  0.9210 $\pm$3.70e-03\\
Adult	    &  0.8589 $\pm$6.57e-03\\
KRFCC	    &  0.9931 $\pm$1.77e-03\\
Breast	    &  0.9222 $\pm$2.73e-03\\
\hline
\end{tabular}
\caption{Two Class problems}
\label{table:corr-woe-2C}
\end{table}

In Table \ref{table:corr-woe-2C}, we observe only Kendall values above 0.82. Kendall's coefficient values can range from 0 to 1. The higher the Kendall's coefficient value, the stronger the association. Usually, Kendall's coefficients of 0.9 or more are usually considered very good.   Kendall's coefficient means also that the appraisers apply essentially the same standard when assessing the samples. With the values shown in this Table, we observe \cite{10.2307/2529310} a minimum of fair agreement to a near perfect agreement  between our Shapley proposition and WoE in terms of ranking of the variable importances\footnote{It would also be interesting to see the correlations of only the most important variables (e.g. the top five), since usually only a few of the most important features are perceptible to humans. However, for lack of space, we do not present this result. We do, however, provide the code for doing so.}.

This good agreement can be understood from two non exclusive perspectives. First, using an averaged naive Bayes model introduces a weight $w_m$ which has a strong influence on the variable importance (the higher the weight, the stronger the influence, for both methods): the variable importance would be mainly influenced by the weights ordering and therefore the same for both methods. Second, it could point out to the fact that the variable-dependent reference terms $-w_m \mathbb{E} \left( log\left(  \frac{P( X_m=x_m | Y_1)}{P( X_m=x_m | Y_0)}  \right)  \right)$ which make the difference between the Shapley value and the WoE are either small or roughly constant in our datasets. How those two perspectives are combined to lead to the good agreement experimentally observed is left for future work.

\subsection{Comparison with Kernel Shap}
\label{expeshap}

Among the libraries able to compute Shapley values, one may find `model oriented' proposals  that can only be used on particular model as for example with tree-based algorithms like random forests and XGBoost (TreeShap \cite{lundberg2018consistent}, FastTreeShap \cite{fasttreeshap}), or model agnostic which can be used with any machine learning algorithm as KernelShap \cite{Lundberg2017}. Here since we did not find a  library dedicated to naive Bayes, we compare our results to the popular  Kernel Shap. In this section we attempt to compare the results obtained, for the Shapley values, with our analytic expression and the results obtained with the  KernelShap library. For a fair comparison, the first point to raise is that the two processes  do not use the same `value function'. Indeed, in our case we use a log odds ratio whereas in KernelShap, when providing the classifier to the library, the value function used is the output of the classifier.\\ 

{\bf On the use of Kernelshap \cite{Lundberg2017}:} The computation time of the library can be very long, even prohibitive. To use the library, the user has to define two datasets: (i) a first dataset, as a knowledge source, which is used to perform the permutation of variable values (ii) a second dataset on which one would like to obtain the Shapley values. The first database is used to compute the Shapley value of the variables for a given example. Given this table and a variable of interest, an example $X_i$,  is modified thanks to the permutation of the others variables. This allows the KernelShap library to create a ``modified table" which contains all the modified versions of this example.

\begin{figure}[!h]
    \centering
    \includegraphics[width=0.4\linewidth]{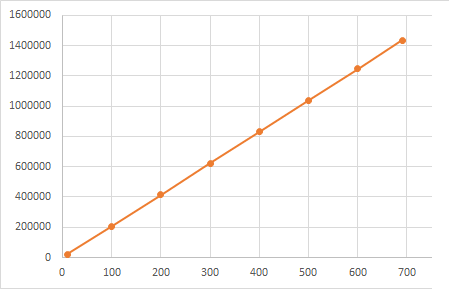}
    \caption{CRX dataset: size of the ``modified table" versus the number of examples in the ``knowledge'' data table.}
    \label{fig:crx}
\end{figure}

To give more intuition about the size of `the modified-example-table' we plot, in Figure \ref{fig:crx}, for the ``CRX" dataset, the size of this table as a function of the number of examples in the `knowledge table', showing  the linear increase that results from a very large table. Then the classifier have to predict its output value for the considered example $X_i$ to compute the Shapley values. For this ``CRX" dataset, which contains 15 inputs variables, the time taken to compute the Kernelshap values for a single example and using all the 690 examples as `knowledge table' is 12.13 seconds\footnote{The characteristics of the used computer are: Intel(R) Core(TM) i7-10875H (No. of threads. 16; 5.10 GHz) RAM:32.0 Go, Windows 10, Python 3.8.6}, so 8370 seconds for the entire dataset (around 2.5 hours for a small dataset). To summarize, the algorithmic complexity of KernelShap is  $O(N_k 2^d)$ where $N_k$ is the number of examples used in the `knowledge table'.

As a consequence, we were not able to obtain a complete result on most datasets (even with a half-day credit) when using the entire dataset. As suggested\footnote{The variance in the results observed in recent publications is due to this constraint.}  by the KernelShap library, in the results below we limit the computation time to a maximum of 2 hours per dataset: (i) the Shapley values are computed only on 1000 (randomly chosen) examples\footnote{It is obvious that for large datasets such as the “adult” the chosen sample of 1000 is statistically insignificant and, as a result, the calculated importance values, computed by KernelShap may not be reliable.} and (ii) the number of examples  in the `knowledge table', $N_k$ \footnote{We start with 50 examples (as a minimum budget) and we increment this number by step of 50 until the credit is reached.}, has been set to the values indicated in the Table~\ref{table:corr-shap-shap-2C} (where the number of examples of the entire dataset is given as a reminder in the brackets).

{\bf On the use of our Shapley proposal -} In contrast, for the analytic Shapley proposed in this paper, the time required to compute the Shapley values is very low (see the discussion in Section \ref{discussion}). Indeed, the algorithmic complexity, for an input variable, is linear in number of parts, intervals or groups of values (see Equation \ref{eqxx}). On the largest dataset used in this paper, the Adult dataset which contains 48842 examples, the time used to compute all the Shapley values for all the variables, all the classes and all the examples is lower than 10 seconds.  This computation time could be further reduced if the $log(P(X|C)$ per variable and per interval (or group values) are precomputed as well as the expectation term of the equation \ref{eqxx}, which is not the case in our experiments.

\begin{table}[!ht]
\centering
\fontsize{9}{9}\selectfont
\begin{tabular}{|l|c|c|c| }
 \hline
Name    &  $N_k$ & Pearson & Kendall \\ \hline 
Twonorm	& 200 (7400)      & 0.9027 & 0.7052 \\
Crx	    & 690 (690)       & 0.9953 & 0.9047 \\
Ionosphere	& 351 (351)   & 0.9974 & 0.8888 \\
Spam	& 200 (4307)      & 0.8829 & 0.7684 \\
Tictactoe	& 958 (958)   & 1.0000 & 1.00 \\
German	& 1000 (1000)      & 0.9974 & 0.9047 \\
Telco	& 1000 (7043)      & 0.9633 & 0.7333 \\
Adult	& 1000 (48842)    & 0.8373 & 0.7692 \\
KRFCC	& 858 (858)       & 0.9993 & 1.00 \\
Breast	& 699 (699)       & 0.9908 & 0.8571 \\
\hline
\end{tabular}
\caption{Correlation between our analytic Shapley and Kernelshap}
\label{table:corr-shap-shap-2C}
\end{table}

\begin{figure}[!h]
    \centering
    \includegraphics[width=0.75\linewidth]{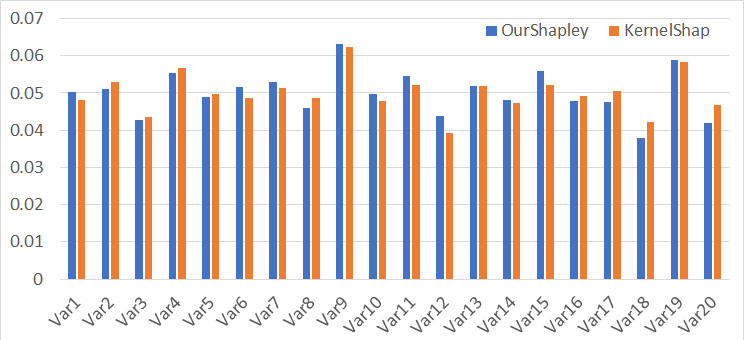}
    \caption{Two Norm dataset: Comparison of our Shapley proposal and KernelShap.}
    \label{fig:twonorm}
\end{figure}

{\bf Results: } The Table \ref{table:corr-shap-shap-2C} gives the correlation between the global Shapley values, defined for each variable as the average on all samples of the absolute values of the local Shapley values. We observe good correlations for both  coefficients. We also give an example of comparison on the TwoNorm dataset in Figure \ref{fig:twonorm} (where we have drawn the normalized global Shapley values), for which the correlations are lowest in the Table \ref{table:corr-shap-shap-2C}. For this data set, the lower Kendall coefficient value is due to the fact that many variables have close Shapley values, resulting in differences in their value ranks. Based on all the results we may conclude that there is a nice agreement between our Shapley proposal and KernelShap on the ten datasets used in this paper.

\section{Conclusion}
\label{sec_conclusion}
In this paper, we have proposed a method for analytically calculating Shapley values in the case of the naive Bayes classifier. This method leverages a new definition of the value function and relies on the independence assumption of the variables conditional on the target to obtain the exact value of the Shapley values, with a linear  algorithmic complexity linear with respect to the number of variables. 
Unlike alternative evaluation/approximation methods, we rely on assumptions that are consistent with the underlying classifier and avoid approximation methods, which are particularly costly in terms of  computation time. We also presented a discussion on the key elements that help to understand the proposal and its behavior.

We compared this Shapley formulation, in an analytic analysis, to another frequently used indicator, the Weight of Evidence (WoE). We also carried out  experiments on ten datasets to compare this proposal with the Weight of Evidence and the KernelShap to highlight similar or dissimilar behaviors. The results show that our Sphaley proposal for the naive Bayes classifier is in  fair agreement with the WoE and with KernelShap's Shapley values, but with a much lower algorithmic complexity, enabling it to be used for very large datasets with extremely reduced computation times.

\clearpage
\bibliographystyle{splncs04}
\bibliography{main}

\end{document}